\documentclass[10pt,twocolumn,letterpaper]{article}

\usepackage{cvpr}
\usepackage{times}
\usepackage{epsfig}
\usepackage{graphicx}
\usepackage{amsmath}
\usepackage{amssymb}
\usepackage{subcaption}
\usepackage{url}

\usepackage[pagebackref=true,breaklinks=true,letterpaper=true,colorlinks,bookmarks=false]{hyperref}

\cvprfinalcopy 

\def\cvprPaperID{****} 

\ifcvprfinal\fi
\begin{document}

\title{Towards Markerless Grasp Capture}

\author{
Samarth Brahmbhatt$^1$, Charles C. Kemp$^1$, and James Hays$^{1,2}$\\
$^1$Institute for Robotics and Intelligent Machines, Georgia Tech $^2$Argo AI\\
{\small \url{samarth.robo@gatech.edu}, \url{charlie.kemp@bme.gatech.edu}, \url{hays@gatech.edu}}
}


\maketitle

\begin{abstract}
Humans excel at grasping objects and manipulating them. Capturing human grasps is important for understanding grasping behavior and reconstructing it realistically in Virtual Reality (VR). However, grasp capture -- capturing the pose of a hand grasping an object, and orienting it w.r.t. the object -- is difficult because of the complexity and diversity of the human hand, and occlusion. Reflective markers and magnetic trackers traditionally used to mitigate this difficulty introduce undesirable artifacts in images and can interfere with natural grasping behavior. We present preliminary work on a completely marker-less algorithm for grasp capture from a video depicting a grasp. We show how recent advances in 2D hand pose estimation can be used with well-established optimization techniques. Uniquely, our algorithm can also capture hand-object contact in detail and integrate it in the grasp capture process. This is work in progress, find more details at \url{https://contactdb.cc.gatech.edu/grasp_capture.html}.
\end{abstract}

\section{Introduction} \label{sec:intro}

Humans, with their complex hands made of soft tissue enveloping a rigid skeletal structure, excel at grasping and manipulating objects. Capturing human grasps of household objects can enable a better understanding of grasping behavior, which can improve a wide variety of VR and human-computer interaction applications. While hand- and object-pose capture and estimation have been studied extensively in isolation, there is a lack of large-scale \textit{grasp capture} datasets and algorithms. We define grasp capture as capture of both the hand and object pose in a scene depicting grasping. Partial occlusion of the object by the hand and vice versa make grasp capture and prediction difficult. As mentioned in Section~\ref{sec:related_work}, the only large scale dataset employs wired magnetic trackers taped to the hand and object~\cite{garcia2018first}. However, this method has the drawback of introducing unwanted artifacts in the RGB images and potentially interfering with natural grasping behavior.

\begin{figure}[h!]
\centering
\includegraphics[width=0.49\textwidth]{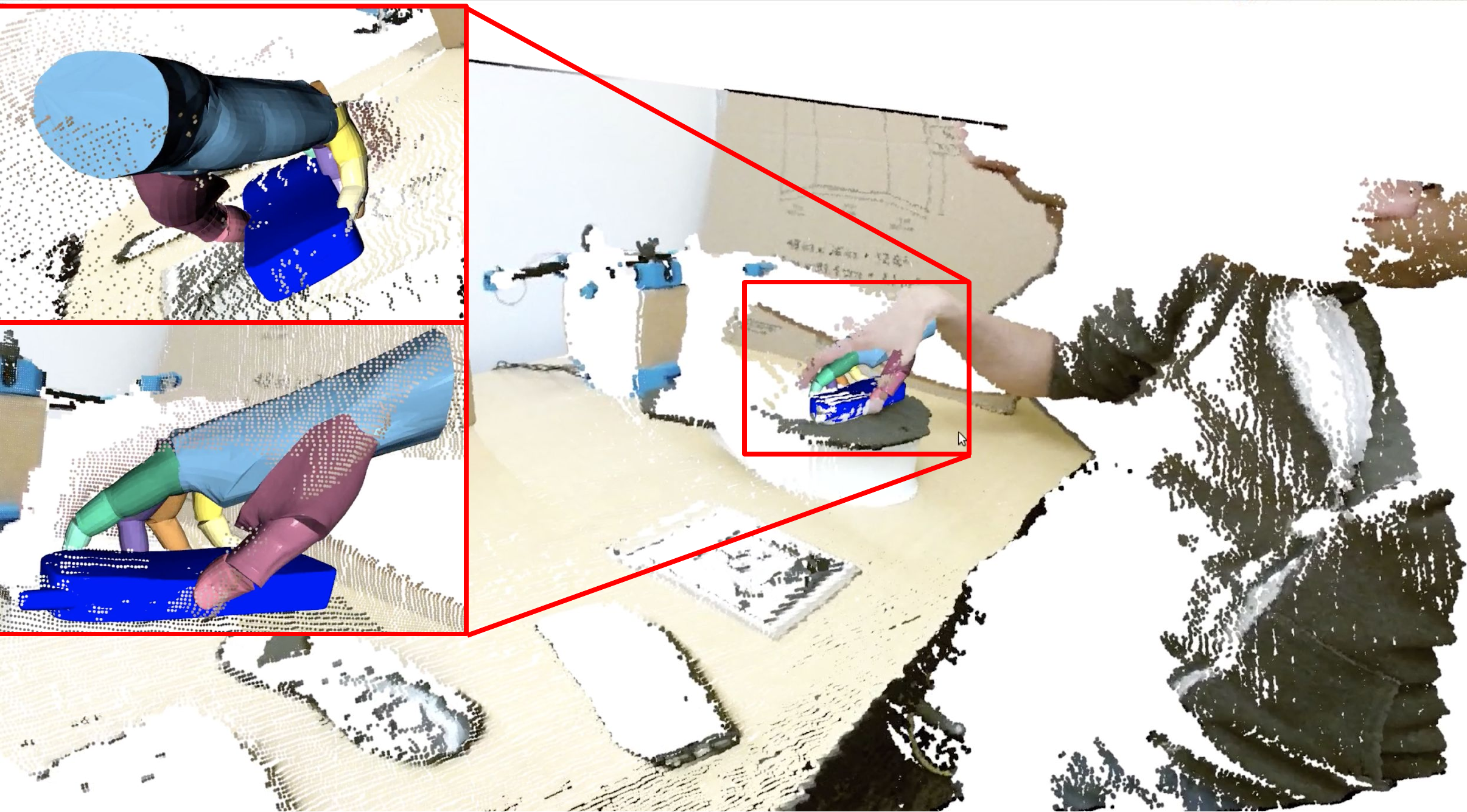}
\caption{Grasp capture for a scene depicting a cellphone being grasped. The multi-colored hand model shows the estimated hand pose and the dark blue model shows the estimated object pose. }
\label{fig:teaser}
\end{figure}

In this paper we focus on capturing the hand pose through the 6-DOF palm pose and 20 joint angles, and the 6-DOF object pose. Since a single image is often not enough to estimate both the hand and object pose, we record a video of a human participant grasping a household object. The participant rotates and translates their hand in 3D space to present the grasp to an RGB-D camera from various perspectives (see Figure~\ref{fig:teaser} for an example frame).

Hand-object contact is either ignored, or enforced without any ground-truth contact observation in traditional grasp capture pipelines. Observing ground truth contact has so far been very difficult, but recently Brahmbhatt et al~\cite{brahmbhatt2019contactdb} created a large scale dataset of detailed hand-object contactmaps through thermal imaging. Different from other grasp capture pipelines, ours can also capture such contactmaps and utilize them improve capture accuracy.
\section{Related Work} \label{sec:related_work}

Hand pose estimation is a highly researched topic, and many datasets are available publicly to train models for hand pose estimation. Hand pose is captured through data gloves~\cite{lin2014grasp, heumer2007grasp}, manually annotated joint locations~\cite{sridhar2013interactive}, magnetic trackers~\cite{wetzler2015rule, Yuan_2017}, or fitting a hand model to depth images after manual initialization~\cite{tompson14tog}. These methods capture only free hands rather than hands grasping objects.

However, as mentioned in Section~\ref{sec:intro}, \textit{grasp capture} also involves capturing the object pose. Relatively few works have addressed this problem. The First Person Hand Action Benchmark~\cite{garcia2018first} is the only large scale real-world dataset capturing both hand and object pose. 3D joint locations and object pose are captured through taped magnetic trackers. In addition to limited working volume (Hasson et al~\cite{hasson19_obman} mention in Section 5.2 that the object poses are imprecise and result in penetration of the hand inside the object by 1.1 cm on average), taping these long wired sensors to the hand introduces artifacts in the RGB images and can potentially interfere with natural grasping behavior.

\subsection{Learning to Predict Aspects of Hand-Object Interaction}
A large number of works estimate the pose of non-grasping hands in a model-based~\cite{tzionas2016capturing, kyriazis2013physically} or model-free~\cite{ye2016spatial, simon2017hand, moon2018v2v} manner. Garcia-Hernando et al~\cite{garcia2018first} note that hand pose estimation in images depicting grasping benefits from including such grasping images in the training dataset. Tekin et al~\cite{tekin2019h+} predict both the hand and object pose by predicting 3D hand joint and object bounding box locations. Hasson et al~\cite{hasson19_obman} predict hand parameters and approximate the object with a predicted genus-0 geometry. Relying on predicted geometry reduces applicability to grasp capture for creating datasets, where object geometry is known in detail. In addition, it is not clear how accurate such algorithms will be on images from a data collection location, which can differ significantly from their training datasets.
\section{Grasp Capture} \label{sec:grasp_capture}

\subsection{Data Collection Protocol} \label{sec:data_protocol}
As mentioned in Section~\ref{sec:intro}, our aim is to capture both the hand pose (6-DOF palm pose and joint angles) and 6-DOF object pose from a video of a human participant grasping a household object. The objects in our experiments are 3D printed at real-life scale using detailed mesh models downloaded from online repositories. Our data collection protocol builds on the protocol from ContactDB~\cite{brahmbhatt2019contactdb}, in which participants hold the object for 5 s and then place it on a turntable, where it is scanned with a calibrated RGB-D-Thermal camera rig. We propose to utilize the object holding time for grasp capture.

\begin{figure}
\centering
\captionsetup[subfigure]{labelformat=empty}
\begin{subfigure}{0.235\textwidth}
	\includegraphics[width=\textwidth]{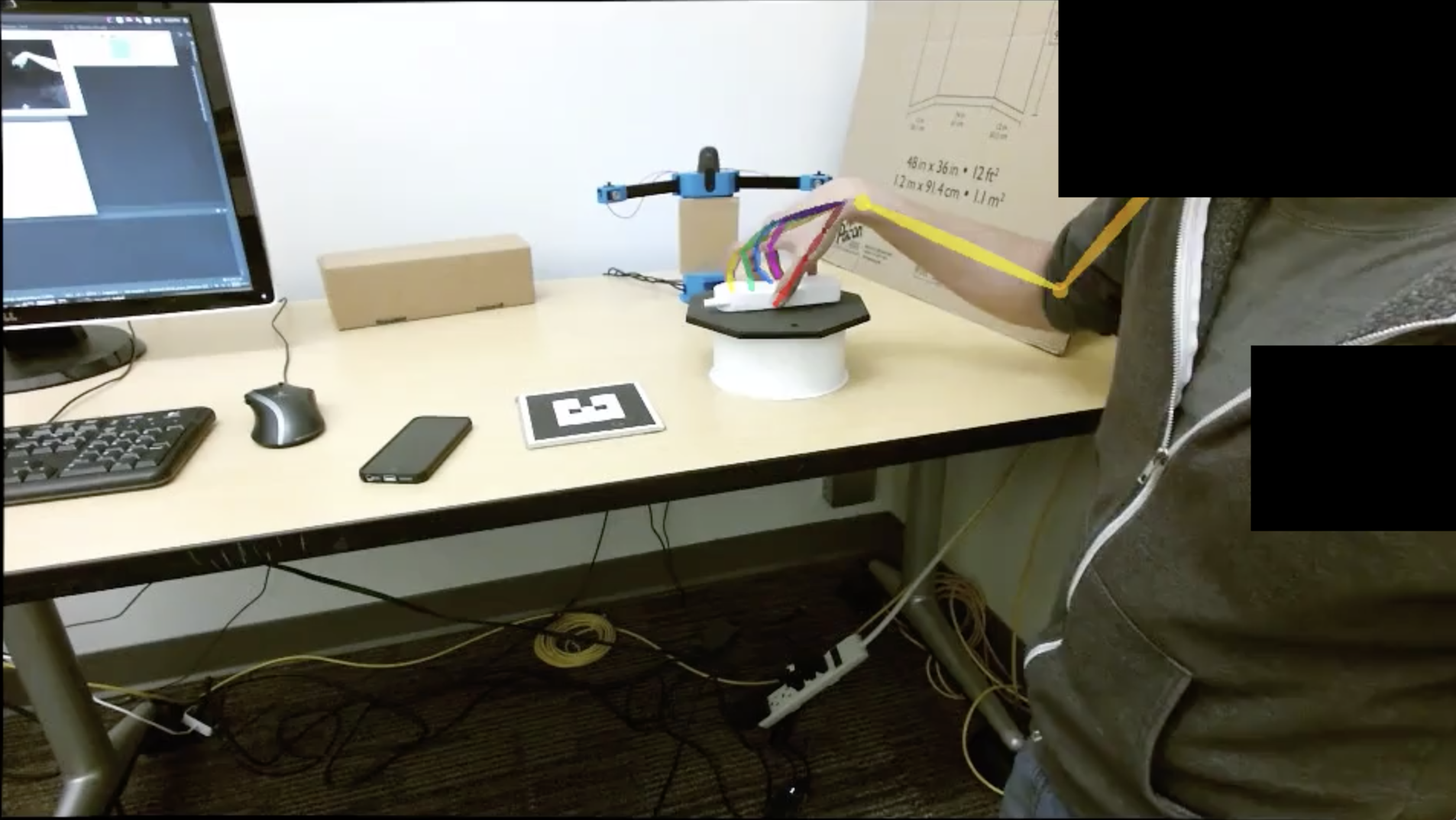}
\end{subfigure}
\begin{subfigure}{0.235\textwidth}
	\includegraphics[width=\textwidth]{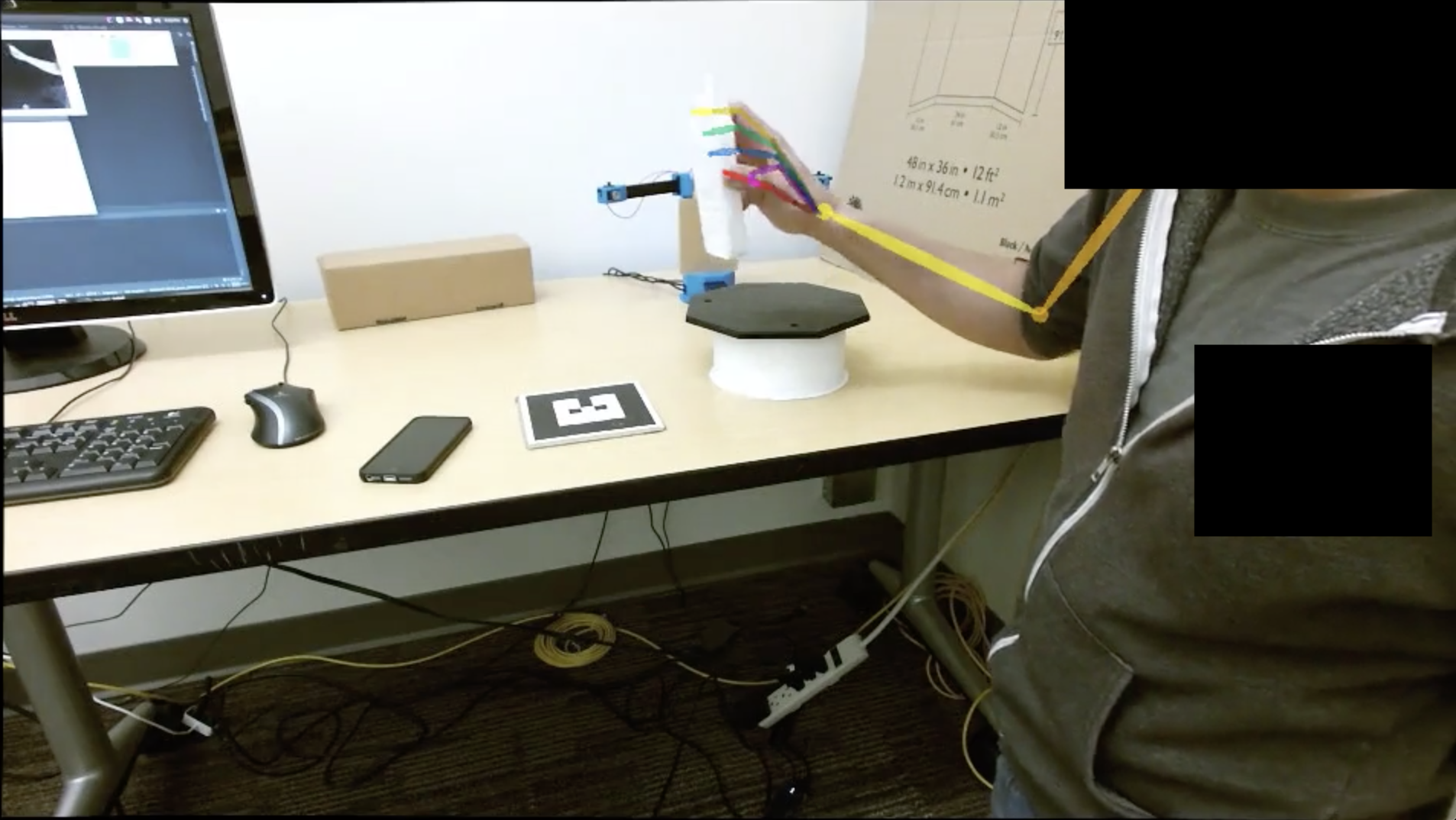}
\end{subfigure}\\
\begin{subfigure}{0.235\textwidth}
	\includegraphics[width=\textwidth]{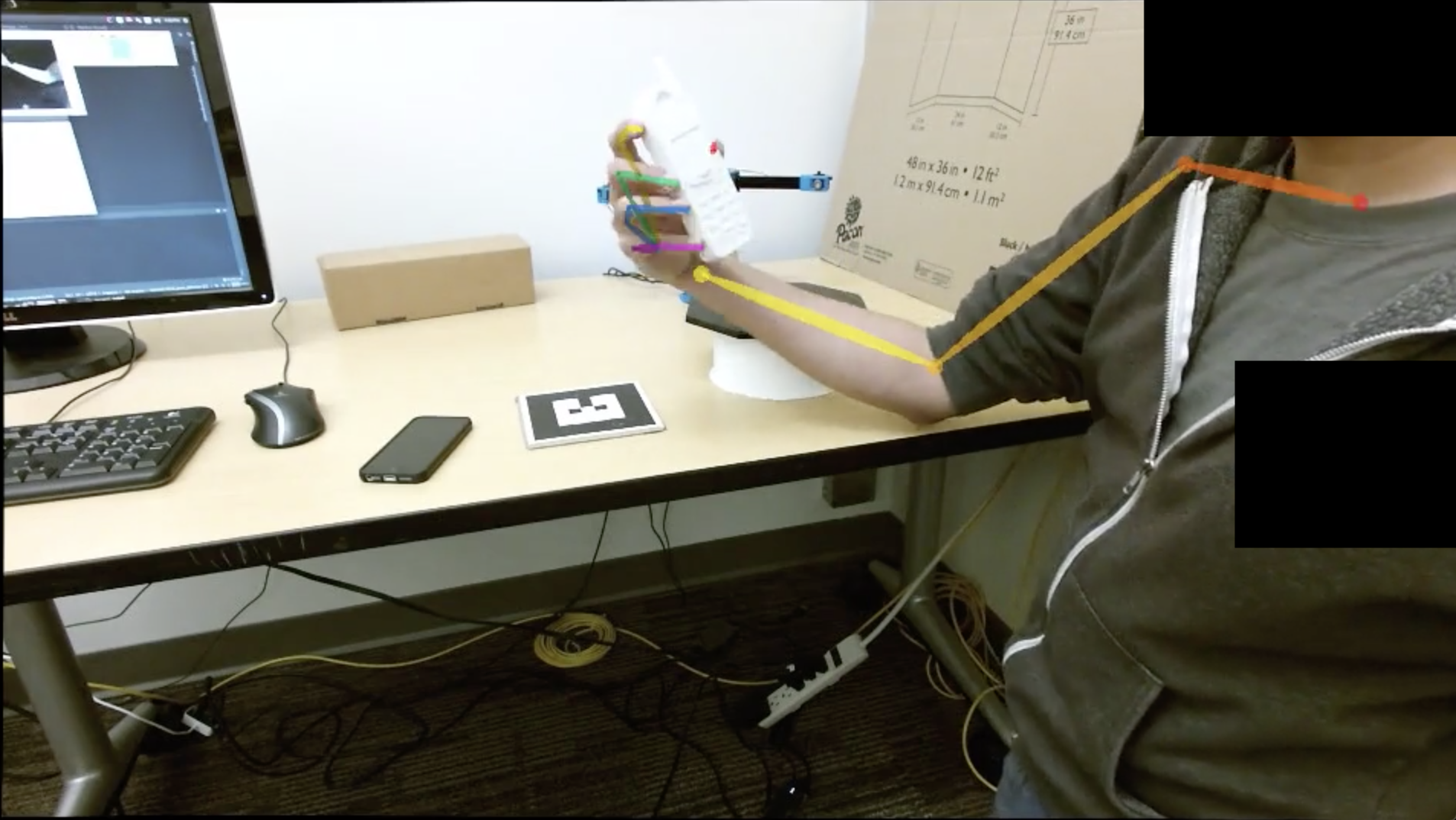}
\end{subfigure}
\begin{subfigure}{0.235\textwidth}
	\includegraphics[width=\textwidth]{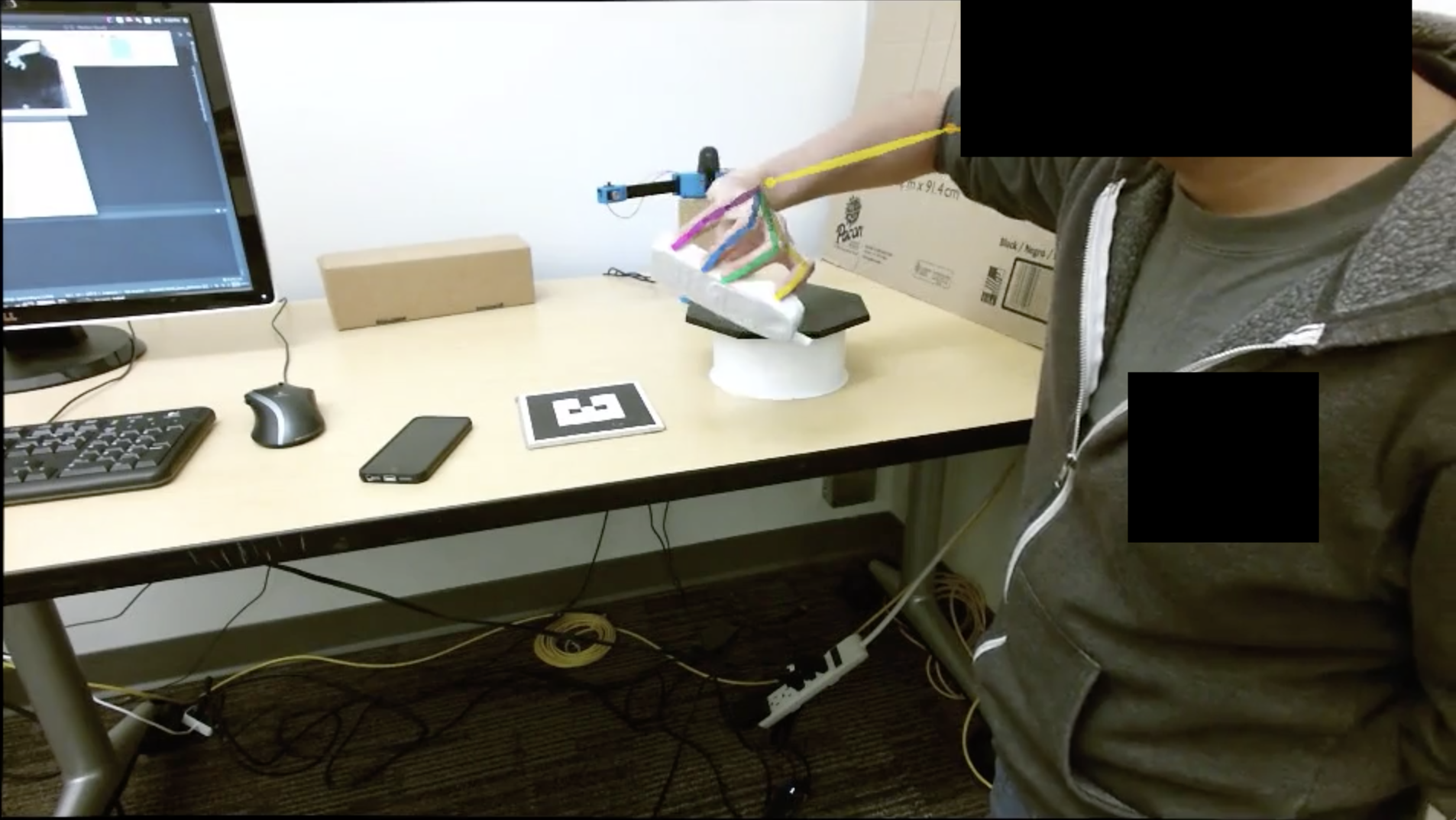}
\end{subfigure}
\caption{OpenPose~\cite{simon2017hand} is used to detect 2D joint locations in the grasp video frames.}
\label{fig:openpose}
\end{figure}

\textbf{Stage 1}: The object is first placed on the turntable, where its 6-DOF pose $^wT_o$ is estimated using the depth camera point-cloud and the known object 3D model.

\begin{figure}
\centering
\includegraphics[width=0.35\textwidth]{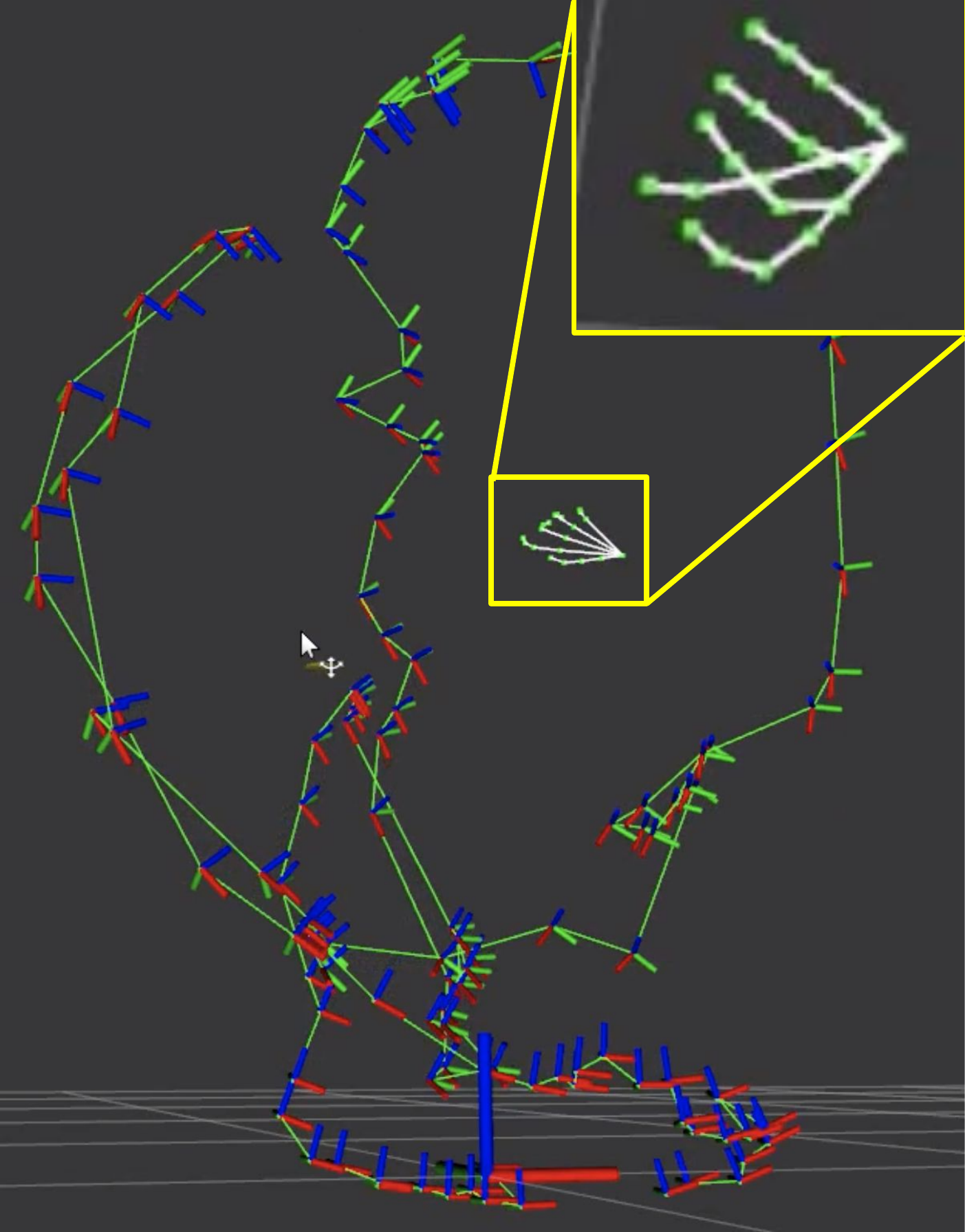}
\caption{Structure from Motion (SfM) is used to recover 3D joint locations (green dots connected by while lines), and virtual camera poses (3D axes with blue pointing along the camera axis). Green lines connect consecutive camera poses.}
\label{fig:sfm}
\end{figure}

\textbf{Stage 2}: The grasp video recording starts when the participant reaches for the object. The participants are instructed to hold their joints steady after a transient phase (termed \textit{grasp adjustment}) in which they pick the object up and settle into a comfortable grasp. Frames of the video after this instant are used to detect 2D hand joints using the OpenPose library~\cite{simon2017hand} (Figure~\ref{fig:openpose}). These 2D detections $\mathbf{x}^{(i)}$ are treated as observations from a mobile virtual camera that is observing a stationary hand (the problem is inverted; in reality we have a stationary camera and a moving hand). A Structure from Motion (SfM) problem is setup using these 2D detections, and optimized using the GTSAM library~\cite{dellaert2012factor} to recover the 3D joint locations $\mathbf{X}$ as well as virtual camera poses $\left\lbrace ^wT_c^{(i)} \right\rbrace_{i=1}^N$, with $^wT_c^{(1)}$ anchored to the origin (Figure~\ref{fig:sfm}). SfM minimizes the following re-projection error:
\begin{equation} \label{eq:sfm}
L \left(\mathbf{X}, \left\lbrace ^wT_c^{(i)} \right\rbrace_{i=1}^N \right) = \sum_{i=1}^N || \mathbf{x}^{(i)} - \pi \left(\mathbf{X}; ^wT_c^{(i)}, K \right) ||_2^2
\end{equation}
where $\pi(\cdot)$ is the camera projection function.

\begin{figure}
\centering
\includegraphics[width=0.35\textwidth]{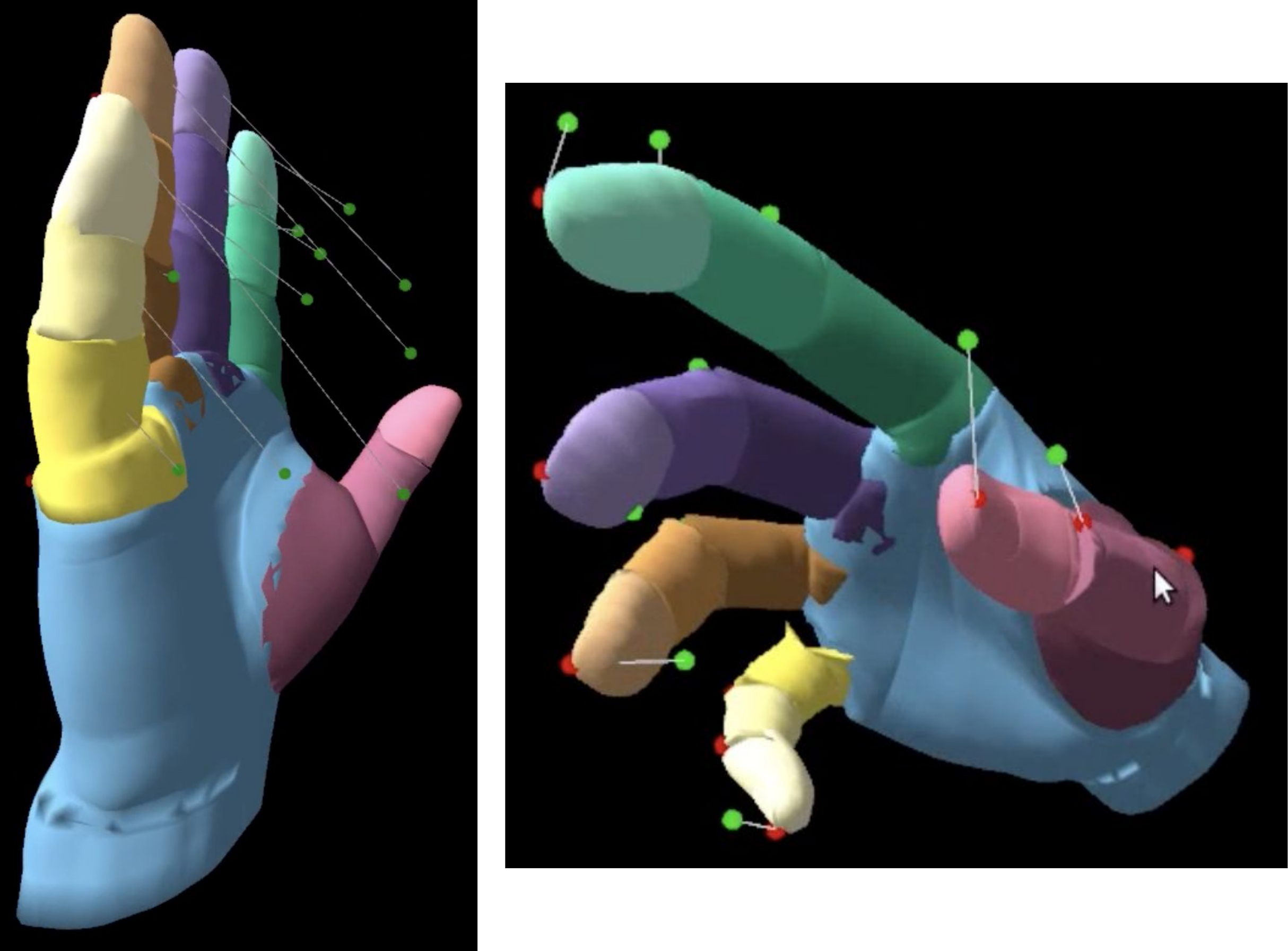}
\caption{Fitting a hand model to the 3D joint locations. Left: Hand transformed by the palm pose. Right: Hand after inverse kinematics fitting. Green dots: target 3D joint locations (from SfM), Red dots: corresponding locations on the hand model.}
\label{fig:hand_fit}
\end{figure}

\textbf{Stage 3}: The hand pose is estimated by fitting a hand model (we use HumanHand20DOF from GraspIt!~\cite{miller2004graspit}) to 3D joint locations $\mathbf{X}$, in two stages (Figure~\ref{fig:hand_fit}): 1) palm pose $^wT_p$ is recovered from the locations of the 6 rigid hand points (wrist base + base of 5 fingers) through the Umeyama transform~\cite{umeyama1991least}, which estimates a 3D similarity matrix, and 2) joint angles are recovered through inverse kinematics after the hand is transformed by $^wT_p$.

\begin{figure}
\centering
\includegraphics[width=0.49\textwidth]{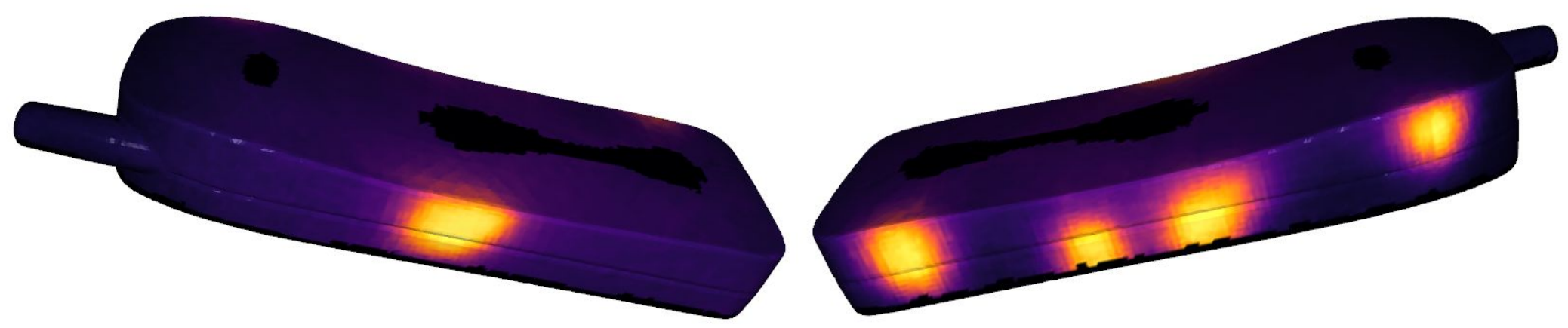}
\caption{The contactmap for the grasp depicted in Figure~\ref{fig:openpose}, captured using ContactDB~\cite{brahmbhatt2019contactdb}.}
\label{fig:contactmap}
\end{figure}

\textbf{Stage 4}: The participant places the object back on the turntable, which starts rotating and the object is scanned by the RGB-D-Thermal camera to construct the contact map according to the ContactDB~\cite{brahmbhatt2019contactdb} protocol (Figure~\ref{fig:contactmap}).

\textbf{Stage 5}: The contact map from Stage 4 can be used to further refine the grasp capture by enforcing the observed contact relation i.e. attracting the closest hand segment to contacted points, repelling it away from non-contacted points, and penalizing intersection of the hand and object (Figure~\ref{fig:teaser}). We follow the grasp optimization stage of the ContactGrasp algorithm~\cite{brahmbhatt2019contactgrasp} to perform this refinement.

The virtual camera poses estimated in Stage 2 can be used to propagate the object and palm pose to all frames of the grasp video: $^cT_o^{(i)} = \left[^wT_c^{(i)} \right]^{-1} {T_{adj}} ^wT_o$, $^cT_p^{(i)} = \left[^wT_c^{(i)} \right]^{-1} {^wT_p}$. Here, $T_{adj}$ is the change in the object pose during grasp adjustment mentioned in Stage 2. To summarize, the proposed algorithm captures the hand- and object-pose for all frames in a video depicting a grasp from various perspectives, without requiring gloves, reflective markers or magnetic trackers.

\subsection{Caveats}
Grasp adjustment (Stage 2), which involves in-hand manipulation~\cite{exner1992hand}, introduces an unknown change $T_{adj}$ in the object pose. We plan to estimate $T_{adj}$ through ICP initialized at $^wT_o$. Another caveat is that OpenPose requires a visible head and shoulder to initialize the hand detector. Since it is not desirable to record the participants' body and face for privacy reasons, we plan to develop a hand detector by skin color segmentation.
\section{Future Work} \label{sec:future_work}

We plan to improve the grasp capture algorithm described in Section~\ref{sec:grasp_capture} in two aspects:
\begin{itemize}
\item Utilizing a more expressive hand model (e.g. MANO~\cite{MANO:SIGGRAPHASIA:2017}) will allow a better fit to individual hand characteristics. Currently, the only identity-dependent parameter in our algorithm the scale estimated during palm fitting (Stage 3). Research in hand modeling has shown~\cite{MANO:SIGGRAPHASIA:2017} that many more parameters are needed to capture the diversity of human hands.
\item Integrating hand-fitting into the SfM problem (Eq.~\ref{eq:sfm}) will reduce the number of stages in the pipeline and make it less brittle. Denoting hand parameters by $\Phi$, we plan to recover hand pose and virtual camera poses jointly by minimizing the following cost function:
\begin{equation}
L \left(\Phi, \left\lbrace ^wT_c^{(i)} \right\rbrace_{i=1}^N \right) = \sum_{i=1}^N || \mathbf{x}^{(i)} - \pi \left( J \left( \Phi \right); ^wT_c^{(i)}, K \right) ||_2^2
\end{equation}
where $J \left( \Phi \right)$ gives the 3D joint locations from $\Phi$.
\end{itemize}
\section{Conclusion} \label{sec:conclusion}

In summary, this paper presents preliminary work on a completely markerless grasp capture algorithm that utilizes well-established geometric optimization techniques and recent advances in 2D hand keypoint detection. In addition to the hand- and object-pose, our algorithm also captures detailed hand-object contact, which is an important component of grasping. We also discuss ways to improve the proposed algorithm. Markerless grasp capture can enable a better understanding of human grasping behavior, and can generate datasets for training models to predict various aspects of grasping like physically plausible hand pose, occurrence of contact at object and hand locations, and potentially even locations and directions of forces being applied to the object~\cite{rogez2015understanding}. These models have applications in VR and human-computer interaction.

{\small
\bibliographystyle{ieee_fullname}
\bibliography{references}
}

\end{document}